# DEEPREFLECS: Deep Learning for Automotive Object Classification with Radar Reflections

Michael Ulrich, Claudius Gläser, and Fabian Timm
*Robert Bosch GmbH*
Stuttgart, Germany
michael.ulrich2@bosch.com

*Abstract*—This paper presents an novel object type classification method for automotive applications which uses deep learning with radar reflections. The method provides object class information such as pedestrian, cyclist, car, or non-obstacle. The method is both powerful and efficient, by using a light-weight deep learning approach on reflection level radar data. It fills the gap between low-performant methods of handcrafted features and high-performant methods with convolutional neural networks. The proposed network exploits the specific characteristics of radar reflection data: It handles unordered lists of arbitrary length as input and it combines both extraction of local and global features. In experiments with real data the proposed network outperforms existing methods of handcrafted or learned features. An ablation study analyzes the impact of the proposed global context layer.

## I. INTRODUCTION

Automotive radar is an important component for advanced driver assistance systems (ADAS) and automated driving (AD). Besides a detection and tracking of objects, the classification of an object's type (e.g. pedestrian, car, non-obstacle) is essential for an accurate interpretation of the vehicle surroundings. However, the classification performance of state-of-the-art automotive radar systems is relatively poor and additional sensors, such as cameras, are often required (see the overview in [1], [2] or the approaches in [3]–[6]).

We see the bottleneck of state-of-the-art object type classifiers in the loss of information through the abstraction of handcrafted features, which are often used, for example in [7]–[10]. While several data-driven high-performance approaches based on spectra or spectrograms were proposed in the literature, e.g. [11]–[13], those approaches are computationally expensive and cannot be used in embedded applications due to high hardware requirements.

In this work, we make the following contributions:

- We propose an object type classifier, which uses radar reflections as input, where each reflection is characterized by its range, radial velocity, radar cross section, and azimuth angle.
- We propose a neural network that classifies the object's type and that can handle a varying number of radar reflections. Still, the size of the neural network is significantly reduced compared to convolutional neural networks.

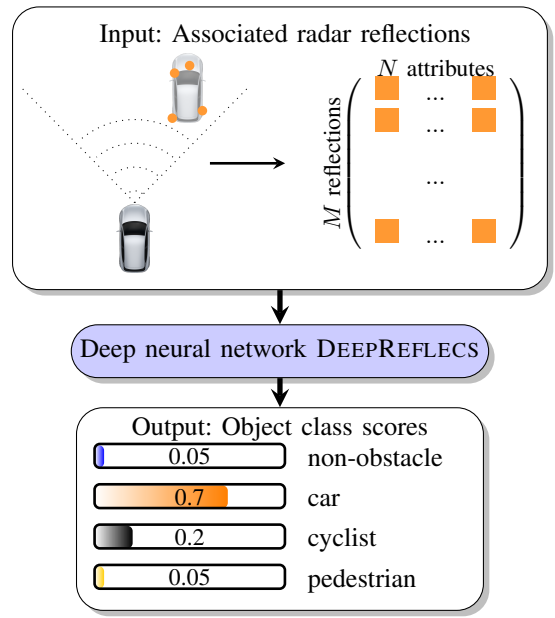

Fig. 1: An overview of the input and output of the proposed object classification.

- We demonstrate the performance of our approach on a real-world dataset with challenging scenarios of stationary objects or slow-moving objects and we compare with different state-of-the-art approaches such as a convolutional neural network and a method with handcrafted features.
- We conduct an ablation study to identify performance variations of our approach.

## II. RELATED WORK

### A. Overview of Methods

Object type classification has been researched for decades. We group the existing solutions in feature-based and data-driven approaches in Tab. I. The feature-based approaches use pre-defined, handcrafted features to obtain a more abstract interpretation of the data. In contrast, data-driven classifiers use the raw data as an input and learn features implicitly during training. The abstraction of features makes feature-based approaches usually computationally less complex, while data-driven approaches often achieve a better performance.

TABLE I: Radar data is used in several applications. Here, we show a brief categorization of some selected approaches. The columns distinguish between a-priori handcrafted and implicitly learned features and the rows distinguish different input representations. This paper fills a previous gap.

| Radar Data Format | Handcrafted Features | Learned Features |
|---|---|---|
| Spectrogram | [10] | [11] |
| Spectrum | [7], [9] | [12], [13], [16]–[18] |
| Radar Reflection Data: | | |
| -Occupancy Grid | [19] | [20] |
| -Single Reflection | [21] | [22], [23] |
| -Reflection List | [24]–[26] | **this work** |

Furthermore, Tab. I distinguishes existing approaches based on the input data. These can be radar spectrograms, radar spectra, or radar reflections.

**Spectrogram as Input:** Spectrograms possess rich information and are used for various application such as human activity classification [11], urban target classification [10] or elderly fall detection [14]. However, spectrograms have to be aggregated over several seconds, which requires a long observability and a good separability of the reflections [15]. Hence, spectrograms are inappropriate for automotive applications with many objects, with high dynamic sensor movements, and low latency requirements.

**Spectrum as Input:** Spectrum-based approaches use a grid representation of the radar data. They can be further grouped by the measured quantity, for example range, Doppler, azimuth angle, or a combination thereof. Many approaches use two-dimensional grids, as in the human vs. robot classification in [16], automotive pedestrian classification [7], or automotive stationary object classification [12]. Few approaches are based on three-dimensional spectra [13] for automotive moving object classification or one-dimensional spectra for pedestrian classification [9], [17], [18].

**Reflections as Input:** Reflection-based approaches usually accumulate data over several measurements. For example, [19] and [20] use occupancy grid maps for automotive object type classification of the stationary environment. Other approaches classify every single reflection, thereby omitting the data association to the objects, e.g. [21] or [22], [23]. Single snapshot measurements without temporal aggregation are used in [24]–[26] for low-latency moving object classification.

Since radar reflections have significantly lower dimension compared to spectrogram and spectrum, but still encode rich information about objects in the environment, we use this input data format and apply a neural network to implicitly extract features. Therefore, our approach fills the gap in current state-of-the-art approaches, see Tab. I.

### B. Why do we use reflections instead of spectra?

The research of spectrum-based data-driven approaches was boosted by the progress in the computer vision community and the transfer of these methods was obvious. The strong focus on the transfer of image processing methods to radar may also explain why most deep learning methods use two-dimensional spectra or spectrograms, although the raw radar data has a higher dimension (range, Doppler, direction-of-arrival). However, we will give some insights why we use radar reflection data instead of spectrogram or spectrum data.

Clearly, radar spectra are the raw representation of our data and many researchers in deep learning argue to pass data as raw as possible to a deep neural network and let the network do the processing. However, we prefer reflection data over spectra for the following reasons:

1) The promise of better performance with raw data is mainly theoretical, holds for unlimited network capacity (number of parameters), and unlimited amount of training data. In practice, this is very expensive or sometimes unlikely to achieve, both for the deployed system (size and processing power for the neural network) and the development (availability of labeled data).
2) Spectrum data for object type classification is expensive. In the deployed radar system, the digital signal processing extracts radar reflections early in the processing chain. Modern automotive radar system use variations of the fast chirp sequence modulation [27], which results in a high amount of raw data. Hence, storing and transmitting spectrum data for object type classification can increase requirements for memory and bandwidth by orders of magnitude.
3) The size (number of parameters) of a neural network to process spectra is usually quite high. This is due to the large size of the input, even when only small parts of the spectrum are used, and given a limited set of labeled data, network overfitting is very likely.
4) Spectrum data is very sensible to modulation and sensor properties. So either an extremely large dataset is required for good generalization performance or each version of the radar system requires its own training dataset, which is infeasible in both cases.

In contrast, reflections require little memory and generalize well over different versions of a radar sensor. Therefore, this work uses radar reflections of a single measurement, where each reflection is characterized by range, radial velocity, radar cross section and azimuth angle—see Fig. 1 (top).

## III. SETUP AND METHODOLOGY

### A. Preprocessing

Reflections proved to contain valuable information of the object type [22], [23], [25], [26]. In contrast to previous literature, we use the reflections, which are already associated to the objects and do not classify each individual reflection. Therefore, we employ a conventional signal processing chain, which is depicted in Fig. 2 and works as follows.

After sampling the analog baseband radar signals, the digital signal processing extracts a list of detected reflections. This is typically done by calculating the range-Doppler spectrum of each channel, detecting reflections using constant false alarm rate (CFAR) thresholding, and direction-of-arrival (DOA) estimation by evaluating the complex amplitudes of the peaks in the range-Doppler spectra of the channels. The data in Sec. V is measured by a commercial radar sensor with a modified chirp sequence modulation for unambiguous range-Doppler estimation and multiple-input multiple-output (MIMO) setup for improved direction-of-arrival estimation. But generally any other modulation and angle estimation principle can be used, as long as there are enough reflections on the object and the measured quantities are accurate.

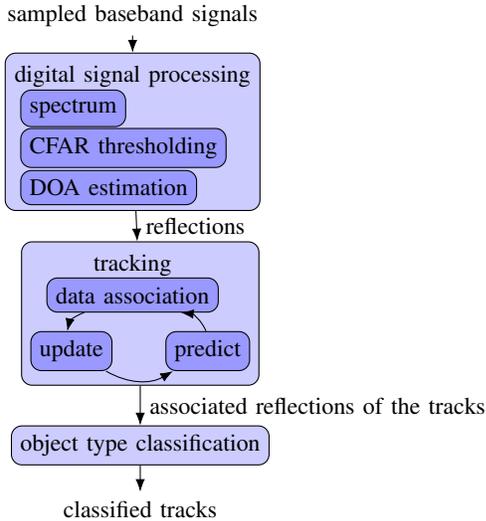

Fig. 2: System architecture of the processing chain. The signal processing extracts reflections from the radar spectra. In the tracking, these reflections are associated to the objects. The associated reflections of each object are used for object type classification.

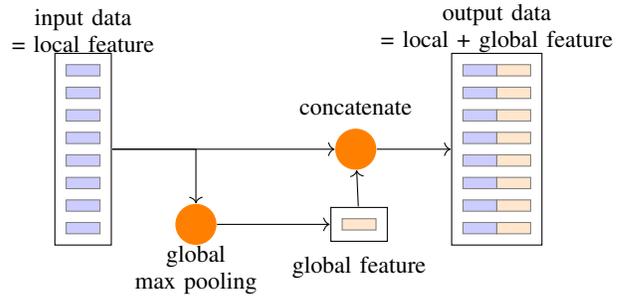

Fig. 3: Block diagram of the global context layer. A pooling layer aggregates the local input features to a global feature vector. This global feature vector is repeated and appended to each individual local input feature.

The radar target list (reflections) is processed by the tracking system. The tracker extracts objects from the target list (one object may cause multiple reflections) by temporal filtering. In each tracking cycle, the tracks of the previous cycle are predicted to the current timestamp, reflections are associated, and the track is updated. The data association of the tracking system estimates the relationship between reflection and object, which is used in the presented object type classifier. The object type classification is processed for each object individually, using only the associated reflections of the corresponding object.

## IV. METHOD

### A. Point processing networks

The two main challenges of processing radar reflections with a deep neural network are the following: The reflection list can have an arbitrary length and the ordering of the reflections in the list is arbitrary. Similar challenges arise when processing point cloud data and we adopt ideas from that domain.

The pioneering work of [28] showed that neural networks benefit from processing point cloud data in its original representation, which is an unordered list. In contrast to previous work, features were extracted from the point cloud directly, without the need for handcrafted features, voxelizaton or projection. This is achieved by applying symmetric functions, for example global max pooling. We briefly revise the key concepts:

- Input representation: The input is a list of points, which are characterized by different features. A point feature could be the Cartesian position of that point, for example. Hence, the input data is one-dimensional (list of points).
- 1-D convolutions: The features of the points in the list must be processed independently to preserve order and size invariance. Hence, a linear transform of the individual features is done using learned weight matrices $\mathbf{M}$ and bias vectors $\underline{b}$ before a non-linear activation function (e.g. ReLU) is applied. All features in one layer are transformed using the same $\mathbf{M}$ and $\underline{b}$. In practice, this is realized using a one-dimensional (1-D) convolution with kernel size one. Although this is not a convolution in the common sense, we adopt this naming to emphasize the weight sharing of $\mathbf{M}$ and $\underline{b}$ for all list entries.
- Global max pooling: While the 1-D convolution processes the information of the points independently, this layer aggregates information over multiple points. One common permutation invariant aggregator is global max pooling. This function selects the maximum of the features in a list per feature.

### B. Global context layer

We propose an architecture based on [28] with an additional layer to add global context information. In the following, we call this type of layer a global context layer. The idea of the global context layer is to combine local and global data during the feature learning, see Fig. 3. For this purpose, we use a permutation invariant aggregator (i.e. global max pooling) on the $M \times N$ input data of the global context layer to generate a $1 \times N$ global feature. The global feature is then concatenated to each individual input list entry (repeated $M$ times).

The global context layer enables the learning of abstract structural features, which was not possible in [28]. The concept of global context layers was already used in graph neural networks [29]. In the proposed method further 1-D convolutions follow the global context layer to learn more abstract features, which capture the structural relationship. For example, consider Cartesian coordinates as input features to a global context layer. The output features will include the Cartesian positions as well as the maximum of all Cartesian positions. Subsequent 1-D convolutional layers with this context information can easily learn the length and width of the object, for example.

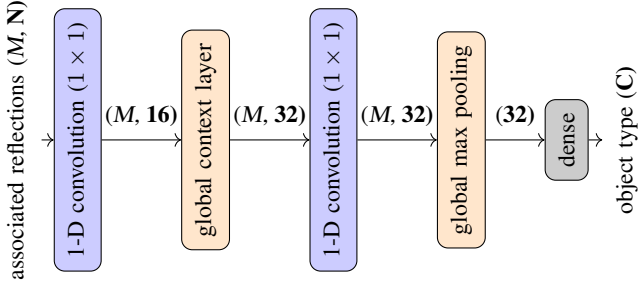

Fig. 4: Architecture of the proposed neural network for object type classification. All layers until the global max pooling are invariant of the order and number of associated reflections. Abstract reflection-wise features can be learned by using the proposed global context layer. The size of the representation is denoted in brackets. For example, ($M$, **16**) is a list of $M$ reflections with 16 features.

TABLE II: Dataset used for the experiments.

| Class | Number of Tracks | Number of Samples |
|---|---|---|
| Car | 574 | 114,825 |
| Pedestrian | 342 | 30,238 |
| Cyclist | 275 | 13,214 |
| Non-Obstacle | 699 | 55,868 |

### C. Radar reflection network

The architecture of the proposed radar reflection network (DEEPREFLECS) is depicted in Fig. 4. The input are the $M$ associated reflections of each object with $N$ features, which were provided by the tracking module, see Fig. 2. Then, a 1-D convolutions with kernel size one follows, which combines the input features to an intermediate representation. The global context layer adds global context to the individual reflection features, see Fig. 3. A further 1-D convolutional layer allows the network to combine the local and global features and consider the relationship of the reflections. Finally, a global max pooling layer converts the list of features to a single feature and a fully connected layer calculates the final object class in a one-hot encoding with $C$ classes.

The global context layer and the global max pooling layer do not have trainable parameters. All other layers use ReLU as nonlinear activation function, except for the last layer, which uses softmax. The size of the trainable tensors **M** and $\underline{b}$ are defined by the input and output tensor shapes of each layer.

The input size of the object type classification is dynamic. To apply the proposed network in practice, the $M$ actual reflections are padded to a list with fixed size (e.g. length 64), such that the input size is larger than the expected input $M$. However, only the first $M$ entries of the padded list are evaluated during the global max pooling operation inside the global context layer or before the final dense layer. By doing so, the padded list entries are ignored and the network can handle any number of input reflections. Furthermore, the 1-D convolutional layers with kernel size 1 as well as the pooling layers are order invariant.

Moreover, filter pruning and quantization techniques [30], [31] can be applied to further reduce the network footprint. Furthermore, the network can be extended for uncertainty estimation [32], which is relevant for safety-critical automotive applications.

### D. Reflection features

The proposed object type classifier uses $N = 5$ reflection features, which are processed in the network. These are:
- The $x$-position of the reflection in the coordinate system of the tracked object.
- The $y$-position of the reflection in the coordinate system of the tracked object.
- The RCS of the reflection.
- The range of the reflection $r$.
- The radial velocity $v_r$ of the reflection, where the ego-motion of the radar is already compensated.

RCS, $r$ and ego-motion compensated $v_r$ are common features for classification, c.f. [22], [23]. The $x$- and $y$-position is converted to the coordinate system of the tracked object, by subtracting the objects $x$- and $y$-position and rotating the reflection positions by the tracked objects heading angle. We use the object coordinate system and $x$-/$y$-position instead of range and angle because the reflection characteristic of an object is approximately constant in Cartesian coordinates and range-independent representations facilitates learning of characteristic patterns.

## V. EXPERIMENTS

We use a dataset of road objects (real measurements), which are relevant in the automotive use-case. The dataset was collected on a test-track and includes tracks from the $C = 4$ classes "pedestrian", "bicycle", "car" and "non-obstacle". In the recording of each track, the host-vehicle with the radar sensor drives towards the object and brakes shortly before it. Consequently, each track consists of multiple samples, which are recorded during the approach. We are using crash-test dummies for the pedestrian and bicycle tracks because such maneuvers would be dangerous to real persons. The class "non-obstacle" contains various overrideable objects which lead to consistent radar reflections, for example a coke can. The size of the dataset is listed in Tab. II. From each track, we use only samples closer than 75 meters. Pedestrians and cyclists have fewer samples per track than cars due to the lack of measurements at larger distance. The cars and non-obstacles are stationary while pedestrians and bicycles are tangentially moving. Hence, this dataset is relatively challenging for object type classification because the radial velocity is not sufficient for accurate classification. The dataset is split into 60% training, 20% validation and 20% test. The split is performed track-wise (samples of one track must not occur in different splits) to avoid overfitting.

TABLE III: Benchmark comparison of the categorical accuracy in % for different classes and methods, evaluated on the self-recorded dataset described in Tab. II.

| Approach | Total | Car | Pedestr. | Cyclist | Non-Obst. |
|---|---|---|---|---|---|
| CRAFTEDFOREST | 90.4 | **96.6** | 78.5 | 47.0 | 92.0 |
| GRIDCNN | 84.3 | 86.6 | 61.9 | 54.7 | 95.4 |
| DEEPREFLECS | **93.4** | 95.8 | **83.9** | **71.4** | **97.6** |

TABLE IV: Ablation results of the categorical accuracy in % for different classes and the impact of the global context layer (GCL) for our approach DEEPREFLECS, evaluated on the self-recorded dataset described in Tab. II.

| Approach | Total | Car | Pedestr. | Cyclist | Non-Obst. |
|---|---|---|---|---|---|
| w/ GCL | **93.45** | **95.75** | 83.89 | **71.42** | **97.58** |
| w/o GCL | 92.20 | 94.86 | **87.25** | 58.89 | 96.35 |
| Diff. | +1.25 | +0.89 | -3.36 | +12.53 | +1.23 |

The proposed network is implemented as depicted in Fig. 4. In total, this network has only 1,284 learnable parameters, which is small for a neural network object type classifier. In the experiments, we compare the proposed reflection processing network with two alternative approaches. The first alternative approach equals [24], abbreviated as CRAFTEDFOREST. It uses handcrafted features such as the second order moments of range and radial velocity. These features are calculated per object and classified using a random forest classifier.

The second alternative applies a 2D convolutional neural network (CNN), abbreviated as GRIDCNN. The input to the CNN is a grid around the tracked object of size $11 \times 11$ ($4m \times 4m$) and with two channels. One channel is the sum of the RCS and the second channel is the average ego-motion compensated radial velocity of all reflections falling into the respective cells. The particular architecture is similar to [12].

More details on the training are given in Appendix A.

## VI. RESULTS

### A. Benchmark

The results of the experiments are described in Tab. III, where we compare the total categorical accuracy as well as the categorical accuracy for each of the four classes. It should be noted that we evaluate the accuracy of single samples, i.e. a temporal smoothing can improve the numbers for all classifiers.

We observe that all classifiers can solve the object type classification task with reasonable total performance. Whereas cars are classified easily due to their size and high RCS, the performance for "pedestrian" and "cyclist" is significantly lower. These classes are often confused with each other and with the "non-obstacle" class.

The performance of the GRIDCNN is worst for most classes among the presented approaches. Although it is a powerful data-driven method, the experiment shows that the grid representation of radar reflection data is inappropriate. One possible explanation could be the high grid sparsity since the radar reflection data only fill parts of the grid.

The CRAFTEDFOREST ranks 2nd in the total accuracy, which we explain by the information loss through the handcrafted features. DEEPREFLECS outperforms the CRAFTEDFOREST by 3% with respect to the total performance and by up to 24% for the most challenging class "cyclist". This demonstrates the impact of a meaningful input representation (radar reflections as a list) as well as the ability to use data-driven feature learning.

In summary, DEEPREFLECS outperforms the other approaches for almost all classes. It further exhibits a computational complexity, which is orders of magnitude lower than CRAFTEDFOREST or GRIDCNN. This is due to the natural representation of radar reflections as a list as well as the introduction of the global context layer.

### B. Ablation Study

The global context layer can capture more complex object properties which a radar sensor cannot measure directly such as width and length, for example. We analyze the impact of the global context layer on the same dataset and quantify the difference in performance. Tab. IV shows the accuracy of our approach DEEPREFLECS with (w/) and without (w/o) the global context layer (GCL). The total performance decreases significantly without the global context layer, from 93.45% to 92.20%. In the minority class "cyclist", the difference is even 12.5%. We explain this by the fact that "cyclist" is particularly difficult to classify because its reflections are similar to "pedestrian" or "non-obstacle". Consequently, the structural relation, which is provided by the global context layer, improves the performance of this class the most.

## VII. CONCLUSION

To summarize, this paper presented a deep learning classifier for automotive radar object type classification. The input representation as a list of reflections is addressed by the use of order- and size-invariant neural network layers, such as 1-D convolution and global max pooling. Further improvements are achieved through the global context layer which combines an extraction of local and global features. The benefits of the proposed approach were demonstrated via experiments on real data. In detail, we could show that the proposed model outperforms previous literature in terms of total classification accuracy by up to 8.8% total performance. For future work, we will investigate whether the global context layer also improves performance for other complex objects such as trucks or trailers.

## APPENDIX A
## TRAINING SETTINGS OF THE EXPERIMENTS

The neural network object type classifiers (GRIDCNN, DEEPREFLECS) are trained in Tensorflow/Python. Each network is trained for $256$ epochs with $1024$ steps and a batch size of $512$. We implemented a learning rate schedule starting at $0.01$ and decreasing to $0.0001$ over the $256$ epochs. For balanced data we increase sample size of "pedestrian" (2x), "cyclist" (2x) and "non-obstacle" (4x) by re-sampling (training dataset only). The validation dataset was used to select the best epoch of a training. The network inputs are normalized to zero-mean and unit-variance. The global max pooling with a masking of unused entries as well as the global context layer are custom implementations, whereas the 1-D convolutions, dense layers, and all layers of the GRIDCNN are already available in Tensorflow.

The GRIDCNN consists of three 2-D convolutional layers with kernel size $3 \times 3$ (channel size 16, 32 and 64), followed by three fully-connected layers with $128$, $32$ and $C = 4$ neurons, respectively. This makes a total of $232{,}628$ learnable parameters for the GRIDCNN. Decreasing the network size in comparison to [12] as well as adding dropout layers was necessary to avoid overfitting, otherwise poor performance was achieved on the test data.

The feature-based object type classifier is trained using the implementation of the sklearn library. The number of decision trees is $100$, otherwise default settings were used. The handcrafted features of [24] are the velocity resolution, the number of reflections per object, the existence of a stationary (low radial velocity) reflection, the average azimuth angle, the average RCS, the average range, the sum of the object extent in $x$ and $y$ Cartesian coordinates, as well as the interval, variance and standard deviation of both range and radial velocity. The sum of the number of nodes in all decision trees of the random forest is $1{,}838{,}940$.